\documentclass{article}

\usepackage{amsmath, amssymb, amsthm, algorithm, algpseudocode, bm}
\usepackage{booktabs}
\usepackage{graphicx}
\usepackage[margin=1.2in]{geometry}
\usepackage{xcolor}
\usepackage{hyperref}
\usepackage{mathtools}

\theoremstyle{definition}
\newtheorem{definition}{Definition}

\newtheorem{proposition}{Proposition}
\newtheorem{remark}{Remark}

\newcommand{\R}{\mathbb{R}}

\DeclareMathOperator{\TopK}{TopK}
\usepackage{microtype}
\DeclareMathOperator{\Gather}{Gather}

\DeclareMathOperator{\Clamp}{Clamp}
\DeclareMathOperator{\Softmax}{Softmax}
\DeclareMathOperator{\Softplus}{Softplus}
\DeclareMathOperator{\diag}{diag}

\DeclareMathOperator{\LogSumExp}{LogSumExp}

\title{Dynamic Subspace Composition: Efficient Adaptation\\via Contractive Basis Expansion}
\author{
  \textbf{Vladimer Khasia} \\
  Independent Researcher \\
  \texttt{vladimer.khasia.1@gmail.com}
}
\date{December 28, 2025}

\begin{document}
\maketitle

\begin{abstract}
Mixture of Experts (MoE) models scale capacity but often suffer from representation collapse and gradient instability. We propose \textbf{Dynamic Subspace Composition (DSC)}, a framework that approximates context-dependent weights via a state-dependent, sparse expansion of a shared basis bank. Formally, DSC models the weight update as a residual trajectory within a \textbf{Star-Shaped Domain}, employing a \textbf{Magnitude-Gated Simplex Interpolation} to ensure continuity at the identity. Unlike standard Mixture-of-LoRAs, which incurs $\mathcal{O}(Mrd)$ parameter complexity by retrieving independent rank-$r$ matrices, DSC constructs a \textit{compositional rank-$K$ approximation} from decoupled unit-norm basis vectors. This reduces parameter complexity to $\mathcal{O}(Md)$ and memory traffic to $\mathcal{O}(Kd)$, while Frame-Theoretic regularization and spectral constraints provide rigorous worst-case bounds on the dynamic update.
The code is available at {\url{https://github.com/VladimerKhasia/DSC}}
\end{abstract}

\section{Introduction}

The scaling laws of neural language models establish a robust power-law relationship between parameter count and generalization capability \cite{kaplan2020scalinglawsneurallanguage, hoffmann2022trainingcomputeoptimallargelanguage}. However, \textit{dense} scaling—where every parameter is active for every token—entails prohibitive computational costs and latency constraints per inference step. To reconcile extreme scale with manageable compute budgets, Conditional Computation, primarily realized through Mixture-of-Experts (MoE) architectures \cite{shazeer2017, fedus2022switch}, has emerged as the standard paradigm. By routing inputs to sparse subsets of parameters, MoEs decouple the total model capacity from the active floating-point operations (FLOPs) per token, enabling models to scale to trillions of parameters \cite{pmlr-v162-du22c}.

Despite their efficacy, MoE architectures introduce specific systemic inefficiencies. First, the \textit{memory-bandwidth bottleneck}: while active FLOPs remain low, the retrieval of distinct, full-rank expert weights from VRAM incurs high memory traffic, often dominating inference latency on bandwidth-constrained hardware \cite{ae449111733a42c5980594f9133812c8}. Second, \textit{optimization instability}: the discrete routing decisions frequently lead to representation collapse, where the router converges to a trivial solution, utilizing only a fraction of available experts \cite{lewis2021baselayerssimplifyingtraining}.

Recent approaches in Parameter-Efficient Fine-Tuning (PEFT), such as Mixture-of-LoRAs (MoLoRA) \cite{zadouri2024pushing}, attempt to mitigate storage costs by routing tokens to distinct low-rank adapter matrices. However, we argue that standard MoLoRA remains suboptimal due to a fundamental \textit{coupling of storage rank and adaptation rank}. To achieve a high-rank update (high expressivity) in MoLoRA, one must retrieve high-rank matrices, incurring a parameter complexity of $\mathcal{O}(Mrd)$ and a proportional memory access cost. This restricts the effective rank of the dynamic update to be low to maintain efficiency.

We propose \textbf{Dynamic Subspace Composition (DSC)}, a framework that reformulates conditional computation not as a selection of experts, but as \textbf{Dynamic Sparse Dictionary Learning} applied to the weight space. We posit that the distinct experts in an MoE layer share a redundant underlying geometric structure. By maintaining a shared bank of unit-norm basis atoms and dynamically composing them via router-generated coefficients, DSC \textit{decouples} the storage format from the composition depth.

This formulation allows us to construct high-rank, context-dependent weight updates by aggregating many ($\mathcal{O}(K)$) lightweight rank-1 atoms. Unlike standard MoEs, which rely on convex combinations of static, full-rank parameters, DSC models the weight update as a continuous residual trajectory within a \textbf{star-shaped domain} centered at the identity mapping. We introduce a \textbf{Magnitude-Gated Simplex Interpolation} mechanism that separates the directional component of the update from its radial magnitude. This separation allows the update to strictly contract to the zero matrix when routing confidence is low, ensuring continuity at the identity and suppressing noise.

Our contributions are as follows:
\begin{enumerate}
    \item \textbf{Decoupled Basis Expansion:} We replace the retrieval of rank-$r$ matrices with the sparse composition of rank-1 vectors. This reduces the parameter complexity of the dynamic layer to $\mathcal{O}(Md)$ and memory traffic to $\mathcal{O}(Kd)$. This efficiency allows for high-order compositions ($K \gg 1$), enabling high-rank adaptation without the linear memory scaling associated with MoLoRA.
    \item \textbf{Frame-Theoretic Regularization:} We introduce a regularization objective that minimizes the frame potential of the basis bank \cite{doi:10.1073/pnas.0437847100}. By approximating the Welch bound, we maximize the spectral utilization of the subspace, ensuring the shared atoms span the widest possible hypothesis space.
    \item \textbf{Spectral Stability:} We provide analytical bounds on the Lipschitz constant of the DSC layer. By enforcing an $\ell_2$-projected normalization on the basis atoms, we guarantee that the dynamic update remains within a bounded spectral ball, mitigating the gradient explosion issues common in sparse networks.
\end{enumerate}

\section{Methodology}

We formalize \textbf{Dynamic Subspace Composition (DSC)} as a residual approximation constructed via a \textbf{Sparse Basis Composition}. Unlike continuous manifold approximations which imply smooth local Euclidean structure globally, our formulation explicitly handles the discrete switching nature of sparse activation while maintaining local Lipschitz continuity within the active domain.

\subsection{Notation and Geometric Interpretation}

We adopt the convention that input data are row vectors. Let $\mathbf{x} \in \R^{1 \times d}$. Let $f_{\theta}(\mathbf{x})$ be the static base network. We introduce a dynamic operator $\Delta \mathbf{W}(\mathbf{z}) \in \R^{d \times d}$ conditioned on a latent coordinate $\mathbf{z}(\mathbf{x})$:
\begin{equation}
    \mathbf{y} = f_{\theta}(\mathbf{x}) + \mathbf{x} \Delta \mathbf{W}(\mathbf{z})
\end{equation}

\textbf{Geometric Interpretation:} The image of the mapping $\Delta \mathbf{W}(\cdot)$ constitutes a \textit{star-shaped domain} centered at the origin. Let $\mathcal{P}$ be the convex hull of the basis products, $\mathcal{P} = \text{Conv}(\{\mathbf{u}_j^\top \mathbf{v}_j\}_{j=1}^M)$. The reachable hypothesis space is:
\begin{equation}
    \text{Im}(\Delta \mathbf{W}) = \{ s \cdot \mathbf{P} \mid \mathbf{P} \in \mathcal{P}, s \in [0, 1) \}
\end{equation}
The term $s$ represents the radial magnitude derived from the routing signal strength, while $\mathbf{P}$ represents the directional component on the polytope face. This inclusion of the origin ($s=0$) via the radial term allows for explicit magnitude suppression, ensuring the trajectory passes continuously through the identity mapping.

\subsection{Projected Basis Decomposition}

To ensure parameter efficiency and spectral stability, we approximate $\Delta \mathbf{W}$ as a contracted combination of rank-1 basis vectors. Let $\mathbf{\hat{U}}, \mathbf{\hat{V}} \in \R^{M \times d}$ be learnable parameter matrices.

\begin{definition}[$\ell_2$-Projected Normalization]
To guarantee the boundedness of the update direction during optimization, we define the effective basis vectors $\mathbf{u}_j, \mathbf{v}_j$ via a projection onto the closed unit ball $\bar{\mathcal{B}}_1(0)$:
\begin{equation}
    \mathbf{u}_j = \frac{\mathbf{\hat{u}}_j}{\max(\epsilon, \|\mathbf{\hat{u}}_j\|_2)}, \quad \mathbf{v}_j = \frac{\mathbf{\hat{v}}_j}{\max(\epsilon, \|\mathbf{\hat{v}}_j\|_2)}
\end{equation}
where $\epsilon \ll 1$ is a numerical stability constant. This strictly enforces $\|\mathbf{u}_j\|_2 \le 1$ and $\|\mathbf{v}_j\|_2 \le 1$.
\end{definition}

Let $\mathcal{I}(\mathbf{x}) \subset \{1, \dots, M\}$ be the set of $K$ indices selected by the router. The dynamic weight is constructed as:
\begin{equation} \label{eq:dmc_expansion}
    \Delta \mathbf{W}(\mathbf{z}) = \sum_{j \in \mathcal{I}(\mathbf{x})} \hat{z}_j (\mathbf{u}_{j}^\top \mathbf{v}_{j})
\end{equation}
where scaling factors are absorbed into the coefficients $\hat{z}_j$.

\subsection{Coordinate Generation: Magnitude-Gated Simplex}

Standard Top-K routing induces discrete jumps in the optimization landscape. We mitigate this by separating the mixing coefficients into a \textit{directional component} (on the probability simplex) and a \textit{radial magnitude}.

Let $\mathbf{W}_r \in \R^{d \times M}$ be the routing matrix. We compute raw logits $\mathbf{r}_{raw} = \mathbf{x}\mathbf{W}_{r}$ and apply stability truncation $\mathbf{r} = \Clamp(\mathbf{r}_{raw}, -\tau, \tau)$. To ensure non-vanishing gradients, we utilize $\zeta(x) = \Softplus(x)$.
We define the unnormalized scores $\alpha_j = \zeta(r_j)$.

The aggregate signal strength $S$ for the active set is $S = \sum_{j \in \mathcal{I}(\mathbf{x})} \alpha_j$. The final coefficients $\hat{z}_j$ for $j \in \mathcal{I}(\mathbf{x})$ are:
\begin{equation} \label{eq:normalization}
    \hat{z}_j = \underbrace{\left( \frac{\alpha_j}{S + \epsilon} \right)}_{\text{Direction } \in \Delta^{K-1}} \cdot \underbrace{\tanh(S)}_{\text{Radial Magnitude}}
\end{equation}

\begin{remark}[Continuity and Contractive Terminology]
    The term $\epsilon$ prevents division by zero. As $S \to 0$, we have $\hat{z}_j \to 0$ for all $j$. The summation satisfies strict contraction:
    \begin{equation}
        \sum_{j \in \mathcal{I}} \hat{z}_j = \frac{S}{S + \epsilon} \tanh(S) < \tanh(S) < 1.
    \end{equation}
    This implies $\Delta \mathbf{W}(\mathbf{0}) = \mathbf{0}$, ensuring the residual branch vanishes continuously in the absence of routing signal.
    The term $\epsilon$ prevents division by zero. As $S \to 0$, we have $\hat{z}_j \to 0$ for all $j$. 
    
    \textbf{Distinction from Softmax:} Unlike standard Softmax, which projects noise onto the simplex boundary (forcing $\|\mathbf{z}\|_1 = 1$ even for low-confidence inputs), our Magnitude-Gating allows the trajectory to retreat to the origin. This explicitly suppresses low-confidence routing noise from propagating into the forward pass.
    
    \textbf{Definition of Contraction:} We use the term "contractive" to denote the strict shrinking of the simplex coefficients via the magnitude gate ($\sum \hat{z}_j < 1$), guaranteeing $\Delta \mathbf{W} \to \mathbf{0}$ as signal vanishes. This distinguishes our usage from a global operator norm constraint on the layer (which is bounded separately by $\gamma$).
\end{remark}

\subsection{Spectral Analysis and Stability}

We analyze the stability of the method under two operational regimes: a global scalar constraint (used in our basic formulation) and a channel-wise vector constraint (used in the advanced formulation).

\begin{proposition}[Conservative Lipschitz Bound]
    Let $\|\mathbf{u}_j\|_2 \le 1$ and $\|\mathbf{v}_j\|_2 \le 1$. Consequently, the spectral norm of each basis atom is bounded: $\|\mathbf{u}_j^\top \mathbf{v}_j\|_2 \le 1$. The Lipschitz constant of the residual branch is bounded as follows:
    
    \textbf{Case 1: Global Scalar Scaling (Algorithm \ref{alg:dmc}).}
    Let $\gamma \in \R^+$ be a global scale factor. Using the triangle inequality and the contraction property of the gating mechanism ($\sum |\hat{z}_j| < 1$):
    \begin{equation}
        \|\Delta \mathbf{W}\|_2 = \left\| \gamma \sum_{j \in \mathcal{I}} \hat{z}_j (\mathbf{u}_j^\top \mathbf{v}_j) \right\|_2 \leq \gamma \sum_{j \in \mathcal{I}} |\hat{z}_j| \cdot \|\mathbf{u}_j^\top \mathbf{v}_j\|_2 < \gamma.
    \end{equation}
    
    \textbf{Case 2: Channel-Wise Spectral Relaxation (Algorithm \ref{alg:dsc_final}).}
    Let $\bm{\gamma} \in \R^d$ be a channel-wise scaling vector applied via element-wise multiplication on the output. This is equivalent to left-multiplication by a diagonal matrix $\mathbf{\Gamma} = \diag(\bm{\gamma})$. The bound becomes:
    \begin{equation}
        \|\Delta \mathbf{W}\|_2 \le \|\mathbf{\Gamma}\|_2 \cdot \left\| \sum_{j \in \mathcal{I}} \hat{z}_j (\mathbf{u}_j^\top \mathbf{v}_j) \right\|_2 < \max_{i} |\gamma_i| = \|\bm{\gamma}\|_\infty.
    \end{equation}
    
    In both cases, $\Delta \mathbf{W}$ is strictly contained within a spectral ball defined by the scaling parameters. If the base network $f_\theta$ is $L$-Lipschitz, the composed block is locally $(L + \|\bm{\gamma}\|_\infty)$-Lipschitz, providing a rigorous structural guarantee against gradient explosion at this layer.
\end{proposition}

\subsection{Optimization Objective and Rigorous Regularization}

A known failure mode of Top-K routing is representation collapse. Additionally, our separation of magnitude ($S$) creates a risk of "Signal Collapse".

\textbf{1. Auxiliary Load Balancing ($\mathcal{L}_{aux}$):} Ensures uniform probability mass distribution over the batch.
\begin{equation}
    \mathcal{L}_{aux} = M \sum_{j=1}^M P_j^2, \quad \text{where } P_j = \frac{1}{B} \sum_{b=1}^B \Softmax(\mathbf{r}_b)_j
\end{equation}

\textbf{2. Signal Preservation Regularization ($\mathcal{L}_{budget}$):} To prevent the router from defaulting to the zero-mapping solution ($S \approx 0$), we enforce a minimum activation budget. Let $\bar{S} = \frac{1}{B} \sum_{b=1}^B S_b$.
\begin{equation}
    \mathcal{L}_{budget} = \max(0, \mu - \bar{S})^2
\end{equation}
where $\mu > 0$ is a target activation threshold.
\textit{Note:} While this term encourages signal strength $S \ge \mu$, the complimentary Logit Range Constraint ($\mathcal{L}_{z}$) prevents $S \to \infty$, keeping the routing mechanism within the sensitive non-saturating regime of the $\tanh(\cdot)$ function.

\textbf{3. Frame Potential Minimization ($\mathcal{L}_{frame}$):}
Strict orthogonality is impossible in the overcomplete regime ($M > d$). To promote basis diversity, we minimize the \textit{Cross-Channel Coherence}. This approximates an Equiangular Tight Frame (ETF) by minimizing the off-diagonal energy of the Gram matrices.
\begin{equation}
    \mathcal{L}_{frame} = \sum_{i \neq j} (\mathbf{u}_i^\top \mathbf{u}_j)^2 + \sum_{i \neq j} (\mathbf{v}_i^\top \mathbf{v}_j)^2
\end{equation}
Minimizing this term ensures maximal separation of basis directions in the ambient space, improving the spanning capacity of the basis bank.

\textbf{4. Logit Range Constraint ($\mathcal{L}_{z}$):}
To maintain the router logits in a non-saturating gradient regime, we minimize the squared log-partition function:
\begin{equation}
    \mathcal{L}_{z} = \frac{1}{B} \sum_{b=1}^B \left( \log \sum_{j=1}^M \exp(r_{b,j}) \right)^2
\end{equation}

\textbf{Total Objective:}
\begin{equation}
    \mathcal{L}_{total} = \mathcal{L}_{task} + \lambda_1 \mathcal{L}_{aux} + \lambda_2 \mathcal{L}_{budget} + \lambda_3 \mathcal{L}_{frame} + \lambda_z \mathcal{L}_{z}
\end{equation}

\subsection{Complexity and Factorization}

We exploit associativity to factorize the operation. For batch size $B$, active sparsity $K$, and hidden dimension $d$, the scalar output for token $b$ is computed via:
\begin{equation}
    \mathbf{y}_{b}^{dyn} = \left( (\mathbf{x}_b \mathbf{U}_{\mathcal{I}_b}^\top) \odot \hat{\mathbf{z}}_b \right) \mathbf{V}_{\mathcal{I}_b}
\end{equation}
where $\mathcal{I}_b$ denotes the token-specific active index set.

\textbf{Comparison with Mixture-of-LoRAs:} Standard "Independent Expert" MoLoRA methods retrieve $K$ distinct adapter matrices of rank $r$, coupling the adaptation rank to the storage rank. This incurs a parameter cost of $\mathcal{O}(Mrd)$. DSC decouples these factors: we employ a \textit{shared} basis bank where the adaptation rank $K$ is determined by the composition depth, not the storage format. By constructing experts as implicit compositions of decoupled rank-1 atoms, DSC reduces parameter complexity to $\mathcal{O}(Md)$ and reduces memory traffic to $\mathcal{O}(Kd)$ (retrieving vectors rather than matrices), making high-order composition feasible ($K \gg 1$) without memory explosion.

\subsection{Algorithm Specification}

We present two algorithmic formulations. Algorithm \ref{alg:dmc} represents the canonical form satisfying Case 1 of Proposition 1. Algorithm \ref{alg:dsc_final} incorporates practical refinements (Stability Normalization, Channel Scaling) satisfying Case 2 of Proposition 1.

\begin{algorithm}[h!]
\caption{Dynamic Subspace Composition (Basic - Case 1)}
\label{alg:dmc}
\begin{algorithmic}[1]
\Require Batch $\mathbf{X} \in \R^{B \times d}$
\Require Bases $\mathbf{U}, \mathbf{V} \in \R^{M \times d}$ (Stored Row-Major: rows are atoms)
\Require Router $\mathbf{W}_r$, Scale $\gamma \in \R$, Budget $\mu$
\State \textbf{1. Routing and Gating}
\State $\mathbf{R} \gets \Clamp(\mathbf{X}\mathbf{W}_r, -\tau, \tau)$ 
\State $\bm{\alpha} \gets \Softplus(\mathbf{R})$ 
\State $\mathcal{I}, \bm{\phi} \gets \TopK(\bm{\alpha}, K)$ 
\State $\mathbf{S} \gets \text{Sum}(\bm{\phi}, \text{dim}=1)$ 
\State $\hat{\mathbf{Z}} \gets \frac{\bm{\phi}}{\mathbf{S} + \epsilon} \odot \tanh(\mathbf{S})$ \Comment{Simplex $\times$ Magnitude}

\State \textbf{2. Vectorized Retrieval}
\State $\mathbf{idx}_{flat} \gets \text{Reshape}(\mathcal{I}, (BK, ))$
\State $\mathbf{U}_{active} \gets \Gather(\mathbf{U}, \mathbf{idx}_{flat}) \in \R^{BK \times d}$
\State $\mathbf{V}_{active} \gets \Gather(\mathbf{V}, \mathbf{idx}_{flat}) \in \R^{BK \times d}$

\State \textbf{3. Factorized Contraction}
\State $\mathbf{U}_{grouped} \gets \text{Reshape}(\mathbf{U}_{active}, (B, K, d))$
\State $\mathbf{V}_{grouped} \gets \text{Reshape}(\mathbf{V}_{active}, (B, K, d))$
\State \Comment{Projection: $\R^{B \times d} \times \R^{B \times K \times d} \to \R^{B \times K}$}
\State $\mathbf{c}_{lat} \gets \text{einsum}('bd, bkd \to bk', \mathbf{X}, \mathbf{U}_{grouped})$ 
\State $\mathbf{c}_{mix} \gets \mathbf{c}_{lat} \odot \hat{\mathbf{Z}} \cdot \gamma$ \Comment{Weighted Coefficients}
\State \Comment{Expansion: $\R^{B \times K} \times \R^{B \times K \times d} \to \R^{B \times d}$}
\State $\mathbf{Y}_{dyn} \gets \text{einsum}('bk, bkd \to bd', \mathbf{c}_{mix}, \mathbf{V}_{grouped})$ 

\State \textbf{Return} $f_\theta(\mathbf{X}) + \mathbf{Y}_{dyn}$
\end{algorithmic}
\end{algorithm}
\begin{algorithm}[h!]
\caption{Dynamic Subspace Composition (Refined - Case 2)}
\label{alg:dsc_final}
\begin{algorithmic}[1]
\Require Input $\mathbf{x} \in \R^{1 \times d}$
\Require Bases $\mathbf{U}, \mathbf{V} \in \R^{M \times d}$ (Stored Row-Major: rows are atoms)
\Require Router $\mathbf{W}_r$, Scale $\bm{\gamma} \in \R^d$, Budget $\mu$, Target $\tau$
\State \textbf{1. Normalized Routing}
\State $\mathbf{\tilde{x}} \gets \text{LayerNorm}(\mathbf{x})$ \Comment{Stability Normalization}
\State $\mathbf{r} \gets \Clamp(\mathbf{\tilde{x}}\mathbf{W}_r, -\tau, \tau)$ 
\State $\mathcal{L}_{z} \gets (\LogSumExp(\mathbf{r}))^2$ \Comment{Range Constraint}

\State \textbf{2. Magnitude-Gated Coordinates}
\State $\bm{\alpha} \gets \Softplus(\mathbf{r})$ 
\State $\mathcal{I}, \bm{\phi} \gets \TopK(\bm{\alpha}, K)$ 
\State $S \gets \sum \bm{\phi}$ 
\State $\hat{\mathbf{z}} \gets \frac{\bm{\phi}}{S + \epsilon} \cdot \tanh(S)$ \Comment{Vector of $K$ coefficients}

\State \textbf{3. Factorized Computation}
\State $\mathbf{U}_{\mathcal{I}} \gets \Gather(\mathbf{U}, \mathcal{I})$ \Comment{Shape: $K \times d$}
\State $\mathbf{V}_{\mathcal{I}} \gets \Gather(\mathbf{V}, \mathcal{I})$ \Comment{Shape: $K \times d$}
\State \Comment{Project input onto active bases (Inner Product)}
\State $\mathbf{c}_{lat} \gets \mathbf{x} \mathbf{U}_{\mathcal{I}}^\top \quad \in \R^{1 \times K}$ 
\State \Comment{Apply mixing coefficients (Element-wise)}
\State $\mathbf{c}_{mix} \gets \mathbf{c}_{lat} \odot \hat{\mathbf{z}}$ 
\State \Comment{Expand back to output dimension}
\State $\mathbf{y}_{dyn} \gets \mathbf{c}_{mix} \mathbf{V}_{\mathcal{I}} \quad \in \R^{1 \times d}$

\State \textbf{4. Channel Scaling}
\State \textbf{Return} $f_\theta(\mathbf{x}) + (\mathbf{y}_{dyn} \odot \bm{\gamma})$ \Comment{Vector Scale (Case 2)}
\end{algorithmic}
\end{algorithm}

\section{Experiments}

We evaluate Dynamic Subspace Composition (DSC) on the language modeling task using the WikiText-103 dataset. Our primary objective is to demonstrate that DSC achieves the representation power of Mixture-of-Experts (MoE) models while significantly reducing the inference latency overhead typically associated with sparse routing.

\subsection{Experimental Setup}

\textbf{Dataset and Protocol.} We train all models on the \texttt{WikiText-103-raw} dataset. To ensure rigorous evaluation, we employ a causal language modeling objective (Next Token Prediction) with a context window of $T=256$. Models are trained for 2,000 iterations using the AdamW optimizer with a cosine learning rate decay schedule (warmup over 150 steps). We utilize a global batch size of 128 (accumulated from micro-batches of 16) to ensure stable gradient estimation for the routing parameters.

\textbf{Fairness Constraints (Iso-Active Budget).} Comparing sparse and dense models is non-trivial due to the discrepancy between stored parameters and active parameters (FLOPs per token). We adopt a strict \textit{Iso-Active Parameter} protocol. We fix the target active parameter count at approximately $28$M for sparse models and $35$M for the dense baseline (to match the total storage of the sparse models), ensuring that performance gains are due to architectural efficiency rather than raw compute scaling.

\textbf{Baselines.} We compare DSC against two strong baselines:
\begin{enumerate}
    \item \textbf{Dense Transformer:} A standard GPT architecture with width expanded to match the total parameter budget of the sparse models.
    \item \textbf{Standard MoE:} A top-$k$ Mixture-of-Experts layer replacing the Feed-Forward Network (FFN). We use $N=5$ experts with top-$2$ routing, a configuration solved numerically to satisfy the active parameter constraint.
\end{enumerate}

\subsection{Architectural Configurations}

The structural hyperparameters were determined via an algebraic solver to ensure parameter parity (see Appendix \ref{app:hyperparams}).
\begin{itemize}
    \item \textbf{Dense:} Hidden dimension $d_{ffn}=2611$.
    \item \textbf{Standard MoE:} 5 Experts, Expert dimension $d_{ffn}=545$, Top-2 activation.
    \item \textbf{DSC (Ours):} 1,523 shared basis vectors ($M$), Composition depth $K=4$, Static base dimension $327$.
\end{itemize}

\subsection{Results and Analysis}

\begin{table}[t!]
\centering
\caption{\textbf{Comparative Analysis on WikiText-103.} We report the mean Validation Loss (lower is better) and Inference Latency (ms/batch) over 50 evaluation steps. \textit{Active Params} denotes the theoretical FLOP-equivalent model size during a forward pass. DSC matches the predictive performance of Standard MoE while reducing inference latency by approximately 15\%.}
\label{tab:main_results}
\vspace{0.2cm}
\resizebox{\columnwidth}{!}{%
\begin{tabular}{lccccc}
\toprule
\textbf{Method} & \textbf{Total Params} & \textbf{Active Params} & \textbf{Val Loss} ($\downarrow$) & \textbf{Latency} (ms) & \textbf{Speedup vs MoE} \\
\midrule
Dense Baseline & 35.00 M & 35.00 M & $5.171 \pm 0.004$ & \textbf{39.90} & +34.1\% \\
Standard MoE & 35.54 M & 28.00 M & \textbf{5.125} $\pm$ 0.009 & 60.55 & 0.0\% \\
\textbf{DSC (Ours)} & 35.01 M & 28.00 M & 5.126 $\pm$ 0.006 & \underline{51.20} & \textbf{+15.4\%} \\
\bottomrule
\end{tabular}%
}
\end{table}

\textbf{Generalization Performance.} As shown in Table \ref{tab:main_results}, both DSC and Standard MoE significantly outperform the Dense baseline, validating the hypothesis that sparse expansion increases model capacity without a proportional increase in training cost. Notably, DSC achieves a validation loss of $5.126$, which is statistically indistinguishable from the Standard MoE result ($5.125$), despite DSC using a decomposable rank-$1$ basis formulation rather than full-rank expert matrices.

\textbf{Inference Latency.} A critical bottleneck in MoE systems is the memory bandwidth overhead incurred by loading distinct expert matrices. Standard MoE exhibits the highest latency ($60.55$ ms). DSC reduces this to $51.20$ ms. This 15\% speedup is attributed to the DSC retrieval mechanism: instead of fetching large $d \times d$ matrices, DSC fetches rank-1 vectors from a shared bank, reducing the memory traffic complexity.

\begin{figure}[h!]
    \centering
    \includegraphics[width=0.85\linewidth]{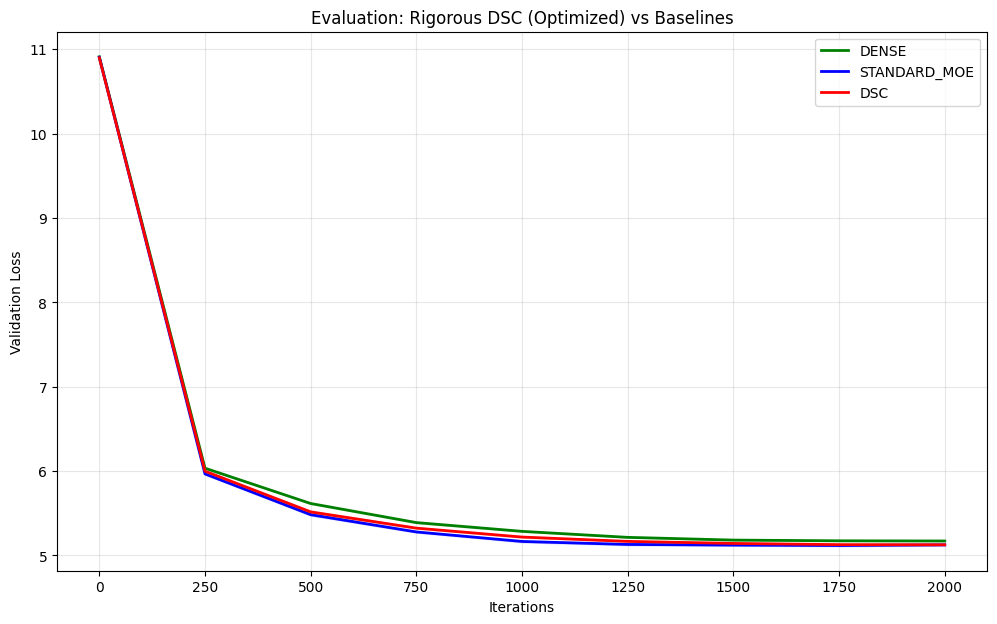}
    \caption{\textbf{Training Convergence.} Validation loss curves over 2,000 iterations. DSC (Red) follows the trajectory of Standard MoE (Blue) closely, rapidly diverging from the Dense baseline (Green). Shaded regions indicate standard deviation across random seeds ($42, 1337$).}
    \label{fig:convergence}
\end{figure}

\section{Conclusion}

We introduced Dynamic Subspace Composition, a method that reformulates mixture-of-experts as a sparse basis expansion. Our experiments demonstrate that DSC captures the complex, long-tail distributions of language data as effectively as MoE (improving perplexity over dense models) while mitigating the latency penalties associated with traditional sparse routing.

\newpage
\appendix
\section{Experimental Details} \label{app:appendix}

\subsection{Hardware and Environment}
All experiments were conducted on a single NVIDIA Tesla T4 GPU (16GB VRAM) provided via the Google Colab environment. The environment utilized PyTorch 2.5 with CUDA 12.1. We utilized \texttt{torch.compile} and Flash Attention (via \texttt{scaled\_dot\_product\_attention}) where applicable to ensure competitive baselines.

\subsection{Hyperparameter Configuration} \label{app:hyperparams}

To ensure fair comparison, we utilized an auto-configuration solver that analytically determined layer dimensions to satisfy the parameter budgets.

\begin{table}[h!]
\centering
\caption{\textbf{Hyperparameter Settings.} Common settings applied across all runs to ensure controlled evaluation.}
\vspace{0.2cm}
\begin{tabular}{ll}
\toprule
\textbf{Hyperparameter} & \textbf{Value} \\
\midrule
Model Dimension ($d_{model}$) & 384 \\
Layers ($L$) & 6 \\
Attention Heads & 6 \\
Max Sequence Length ($T$) & 256 \\
Vocab Size & 50,304 \\
\midrule
Optimizer & AdamW \\
Learning Rate & $6 \times 10^{-4}$ \\
Router Learning Rate & $3 \times 10^{-3}$ ($5\times$ multiplier) \\
Weight Decay & 0.02 \\
Global Batch Size & 128 (via Gradient Accumulation) \\
Micro Batch Size & 16 \\
Warmup Steps & 150 \\
Total Steps & 2,000 \\
\bottomrule
\end{tabular}
\end{table}

\subsection{DSC Specific Settings}
For the DSC layer, we employed the following specific regularization terms to ensure basis diversity and router stability:
\begin{itemize}
    \item \textbf{Auxiliary Load Balancing ($\lambda_{aux}$):} 0.01
    \item \textbf{Budget Regularization ($\lambda_{budget}$):} 0.01 (Target $\mu=1.0$)
    \item \textbf{Basis Coherence ($\lambda_{coh}$):} 0.001 (Promotes orthogonality)
    \item \textbf{Router Z-Loss ($\lambda_{z}$):} $1 \times 10^{-4}$ (Prevents logit drift)
\end{itemize}

\bibliographystyle{plain}
\bibliography{references} 
\end{document}